\documentclass[journal]{IEEEtran}
\ifCLASSINFOpdf
\usepackage{cite}
\usepackage{lscape}
\usepackage{textcomp}
\usepackage{multirow}
\usepackage[graphicx]{realboxes}
\usepackage{adjustbox}
\usepackage{tabularx}
\usepackage[normalem]{ulem}
\usepackage{tablefootnote}
\usepackage{xcolor}
\usepackage{verbatim}
\usepackage{algorithmic}
\usepackage{amsmath,amssymb,amsfonts}
\usepackage{amsthm}
\usepackage{lscape}
\usepackage{hyperref}
\usepackage{soul}
\usepackage[ruled, vlined, linesnumbered]{algorithm2e}

\newcommand{\blu}{\color{blue}}
\newcommand{\blk}{\color{black}}

\def\BibTeX{{\rm B\kern-.05em{\sc i\kern-.025em b}\kern-.08em
    T\kern-.1667em\lower.7ex\hbox{E}\kern-.125emX}}

\begin{document}

\renewcommand{\algorithmicrequire}{\textbf{Input:}}
\renewcommand{\algorithmicensure}{\textbf{Output:}}

\title{DeepHealthNet: Adolescent Obesity Prediction System Based on a Deep Learning Framework}

\author{Ji-Hoon Jeong,~\IEEEmembership{Associate Member, IEEE}, In-Gyu Lee, Sung-Kyung Kim, Tae-Eui Kam, \\ Seong-Whan Lee*,~\IEEEmembership{Fellow, IEEE}, and Euijong Lee*
\blu{
\thanks{Ji-Hoon Jeong, In-Gyu Lee, and Euijong Lee are affiliated with the School of Computer Science, Chungbuk National University, Chungbuk 28644, South Korea. E-mail: jh.jeong@chungbuk.ac.kr, ingyu.lee@chungbuk.ac.kr, kongjjagae@cbnu.ac.kr}
\thanks{Sung-Kyung Kim is affiliated with the Injewelme Corporation, Seoul 02841, South Korea. E-mail: skkim@injewelme.com}
\thanks{Tae-Eui Kam and Seong-Whan Lee are affiliated with the Department of Artificial Intelligence, Korea University, Seoul 02841, South Korea. E-mail: kamte@korea.ac.kr, sw.lee@korea.ac.kr}
}}\blk

\markboth{}
{}

\maketitle
\begin{abstract}
Childhood and adolescent obesity rates are a global concern because obesity is associated with chronic diseases and long-term health risks. Artificial intelligence technology has emerged as a promising solution to accurately predict obesity rates and provide personalized feedback to adolescents. This study emphasizes the importance of early identification and prevention of obesity-related health issues. Factors such as height, weight, waist circumference, calorie intake, physical activity levels, and other relevant health information need to be considered for developing robust algorithms for obesity rate prediction and delivering personalized feedback. Hence, by collecting health datasets from 321 adolescents, we proposed an adolescent obesity prediction system that provides personalized predictions and assists individuals in making informed health decisions. Our proposed deep learning framework, DeepHealthNet, effectively trains the model using data augmentation techniques, even when daily health data are limited, resulting in improved prediction accuracy (acc: 0.8842). Additionally, the study revealed variations in the prediction of the obesity rate between boys (acc: 0.9320) and girls (acc: 0.9163), allowing the identification of disparities and the determination of the optimal time to provide feedback. The proposed system shows significant potential in effectively addressing childhood and adolescent obesity.
\end{abstract}\


\IEEEpeerreviewmaketitle
\section{Introduction}
\IEEEPARstart {C}{hildhood} and adolescent obesity rates have become a growing concern worldwide in recent years. According to the World Health Organization (WHO), the number of overweight children and adolescents aged 5--19 years has increased from 32 million in 1990 to 42 million in 2013, globally \cite{abarca2017worldwide, ferreras2023systematic, degregory2018review}. Obesity in childhood and adolescence is associated with various health problems, such as cardiovascular disease, type 2 diabetes, and musculoskeletal disorders, and can lead to a higher risk of obesity and related health problems in adulthood \cite{caprio2020childhood}.

Predicting obesity rates is crucial because early identification of individuals at risk can help prevent and manage obesity-related health problems \cite{ahmed2009childhood, hammond2019predicting, pang2019understanding}. Artificial intelligence (AI) technology has been widely used recently to implement digital healthcare using various approaches \cite{torner2019multipurpose}, such as implementing smart homes using Internet of Things \cite{saadeh2019patient}, building multimodality interfaces for real-world applications \cite{ang2016eeg, jeong2020brain}, and providing feedback to users using biomedical sources \cite{zhao2022uda, jeong2022real}. Additionally, AI technology is potentially beneficial to this field through its accurate and personalized predictions of obesity rates for adolescents. AI algorithms can be trained on large health information datasets and provide tailored feedback to individuals, allowing them to make informed health decisions \cite{yu2018artificial, lee2021decoding}.

\begin{figure*}[t!]
\begin{center}
\includegraphics[width=0.85\textwidth]{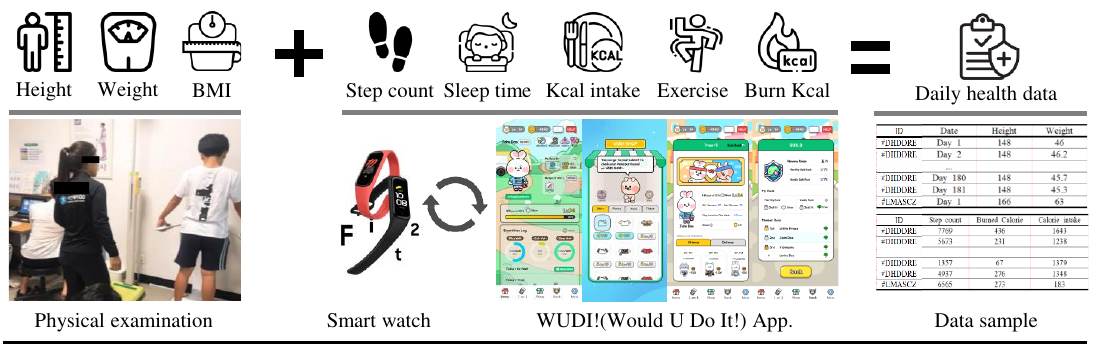}
\caption{Data configuration for acquiring daily health data (height, weight, body mass index (BMI), step count, sleep time, kcal intake, exercise, and burned kcal). Data were collected by physical examination, a smartwatch, and the WUDI! application from 321 participants.}
\end{center}
\end{figure*}

The rising prevalence of childhood and adolescent obesity has become a significant public health concern. Recently, there has been an increasing focus on developing accurate and effective approaches for predicting obesity rates in children and adolescents \cite{bastida2023promoting}. A key factor in this effort is the collection and analysis of relevant health data \cite{yang2020homecare}. Various health-related data, such as height, weight, waist circumference, calorie intake, physical activity levels, and other relevant health information, must be collected to accurately predict the obesity rates of children and adolescents \cite{huffman2010parenthood}. These data are essential in developing accurate algorithms to predict obesity rates and provide personalized feedback to individuals. The availability of such datasets can enable researchers and healthcare providers to develop more effective interventions and prevention strategies for childhood and adolescent obesity \cite{ mondal2023predicting}. However, there are various challenges in collecting and analyzing these data, such as ensuring data quality, protecting privacy, and addressing ethical considerations.

Several studies have examined the utilization of AI technology for predicting obesity rates among children and adolescents. Gupta et al. \cite{gupta2022obesity} trained a general long short-term memory (LSTM) network using both static and dynamic electronic health record data over a period of 1--3 years to predict obesity for individuals ranging between 3--20 years. On average, they achieved an area under the curve score of 0.88. Mondal et al. \cite{mondal2023predicting} employed a machine learning (ML) classifier to categorize individuals into three groups based on childhood health maintenance data: normal weight, overweight, and obese. The experimental outcomes demonstrated the classification accuracies of 89\%, 77\%, and 89\% for the three respective scenarios. Furthermore, Cheng et al. \cite{cheng2022predicting} utilized the Obesity Prediction in Early Life (OPEL) database as their dataset. After pre-processing the data, they divided the children who had clinical visits 2, 3, 5, and 8 times between ages 0 and 4 into male and female groups. They trained an LSTM model and obtained a mean absolute error of 0.98 and a lasso regression value of 0.72.

The primary objective of this study was to investigate the application of AI technology in predicting obesity rates in adolescents. The study aimed to assess the capabilities of AI algorithms to forecast obesity rates and provide personalized feedback to individuals. Additionally, the study explored the importance of data collection, a crucial aspect in predicting obesity rates in adolescents. The types of data required for accurate predictions, the challenges involved in collecting and analyzing such data, and the potential advantages of data-driven approaches in tackling childhood and adolescent obesity were examined. 

The study also proposes a prediction system that uses an AI model to predict the likelihood of obesity in adolescents proactively and offers tailored feedback based on these predictions. The proposed model, DeepHealthNet, outperformed other comparable models in accurately predicting obesity rates. DeepHealthNet effectively employs data augmentation techniques to train deep learning models, even when the available daily health data is limited, resulting in improved predictive performance. Furthermore, the study investigated the variations in obesity rate predictions between boys and girls, aiming to reveal any disparities and determine the optimal timing for delivering feedback. The implementation of the proposed system has significant potential to address the problem of childhood and adolescent obesity.

\begin{table*}[t!]
\caption{Data Examples with respect to Obesity for Adolescents. Factors of Daily Health Data: Gender, Grade, Birth, Height, Weight, BMI, Step Count, Total Sleep Time, Calorie Intake, Amount of Food Intake, Exercise, Exercise Time, Burned Calories}
\renewcommand{\arraystretch}{1.2}
\resizebox{\textwidth}{!}{%
\begin{tabular}{c|lccccccccccccc}
\hline
INDEX                                             & \multicolumn{1}{c}{Subjects}                                          & Gender                                                   & Grade                                             & Birth                                             & \multicolumn{1}{c}{Height}                                            & \multicolumn{1}{c}{Weight}                                            & \multicolumn{1}{c}{BMI}                                               & \multicolumn{1}{c}{Step count}                                        & \multicolumn{1}{c}{Total sleep time}                                  & \multicolumn{1}{c}{Calorie intake}                                        & Amount of food intake                                                                                           & Exercise                                          & Exercise time                                     & \multicolumn{1}{c}{Burned calories}                                     \\ \hline
UNIT                                              & \multicolumn{1}{c}{ID}                                                & \begin{tabular}[c]{@{}c@{}}1: Boy\\ 2: Girl\end{tabular} & 4$\sim$6                                         & YY-MM                                          & \multicolumn{1}{c}{cm}                                                & \multicolumn{1}{c}{kg}                                                & \multicolumn{1}{c}{$kg/m^2$}                                             & \multicolumn{1}{c}{Per day}                                           & \multicolumn{1}{c}{Minute}                                            & \multicolumn{1}{c}{kcal}                                              & \multicolumn{1}{l}{\begin{tabular}[c]{@{}l@{}}1:   Light \\ 2: Moderate \\ 3: Heavy\end{tabular}} & Type                                              & Minute                                            & \multicolumn{1}{c}{kcal}                                              \\ \hline
1                                                 & \#DHDDRE                                                              & 2                                                        & 6                                                 & 2010-07                                        & 153.0                                                                 & 41.0                                                                  & 17.51                                                                 & 4651                                                                  & 511                                                                   &1063                                                                  & 1                                                                                                 & 5                                                 & 15                                                & 40.16                                                                 \\
2                                                 & \#IGMKBO                                                              & 2                                                        & 6                                                 & 2010-06                                        & 152.2                                                                 & 38.0                                                                  & 16.40                                                                 & 2262                                                                  & 548                                                                   & 175                                                                   & 2                                                                                                 & 5                                                 & 47                                                & 126.06                                                                \\
3                                                 & \#EMASCZ                                                              & 2                                                        & 6                                                 & 2010-07                                      & 163.0                                                                 & 47.0                                                                  & 17.69                                                                 & 2454                                                                  & 325                                                                   & 544                                                                   & 2                                                                                                 & 5                                                 & 14                                                & 33.33                                                                 \\
4                                                 & \#EXVWOM                                                              & 2                                                        & 6                                                 & 2010-11                                        & 143.5                                                                 & 36.5                                                                  & 17.73                                                                 & 4075                                                                  & 580                                                                     & 179                                                                   & 2                                                                                                 & 10                                                & 5                                                 & 21.13                                                                 \\
5                                                 & \#QMWDAI                                                              & 2                                                        & 6                                                 & 2010-08                                        & 147.0                                                                 & 38.0                                                                  & 17.59                                                                 & 520                                                                    & 332                                                                   & 174                                                                   & 2                                                                                                 & 5                                                 & 15                                                & 35.18                                                                 \\
\hline
\end{tabular}}
\end{table*}

\section{Materials and Methods}
\subsection{Participants}
Initially, 321 participants (aged 10--12 years, 133 males and 188 females), who were students of the same elementary school in Seoul, participated in the use of the WUDI! mobile application. Of these 321 participants, 187 (75 males and 112 females) officially underwent body measurements by the Korea Sports Promotion Foundation (KSPO) for the experiment, immediately before and after the experimental period. The overall experimental protocols and environments were reviewed and approved by the Institutional Review Board of Chungbuk National University (CBNU-202308-HR-0196).

The participants were healthy, without any neurophysiological anomalies, musculoskeletal disorders, or growth hormone deficiencies. Before the experiments, they were briefed on how to use the mobile application and smartwatch and to sync the data via an animated tutorial. Parents and school staff, including principal and class teachers, were informed of the experimental protocols, paradigms, and purpose. After ensuring that the parents and guardians of the participating students had understood the information, their written consent was obtained according to the Personal Information Protection Act of Korea, and their signature was obtained on a form that specified their consent to the anonymous public release of data. The physical and mental states of the participating students were evaluated to compare the effect of the application used on individual states. Additionally, each participant was required to maintain their normal daily routines and to be in a normal health condition during the experiments.

After submitting the signed consent form, each participant received a Samsung Galaxy Fit 2 smartwatch, prepared for the experiment to acquire detailed data on their activity and sleep for better accuracy and further analysis. As the application developer, Injewelme Co., Ltd., has officially partnered with Samsung Health, the data collected by the Samsung device were automatically synchronized from Samsung Health to the provided application WUDI! (Would U DO It!) for this experiment, via application integration (Fig. 1)

\subsection{Experimental Protocols for Data Acquisition}
The body measurements of the participants were obtained twice. For this experiment, KSPO supported by providing measuring tools and two staff members in charge of the Songpa Fitness Certification Center to the participating elementary school immediately before (3rd week of July 2021) and after (5$^{th}$ week of September 2021) the experiment. Offline measurements consisted of three parameters: 1) height, 2) weight, and 3) waist size, to confirm and compare the body changes of the individuals objectively. A simple survey was administered to the students asking about the kind of gift they would like to receive as a surprise. The purpose of this survey was to prepare for a later intervention to increase participation and attract more attention to WUDI!.

Furthermore, prior to the experiment, each participant and their parents had to log in to WUDI!. First, the parents needed to agree to the terms of service and input the basic information of their child including height and weight. As consent for service use is required from parents of minors under 14 years of age as per Korean law, the participants could obtain the invitation code to activate their use of the WUDI! application only after the parents had completed the registration properly. When the students installed WUDI!, they had to input the invitation code received via MMS and then watch an animated tutorial, which demonstrated how to use the mobile application properly and sync the smartwatch.

\begin{figure*}[t!]
\begin{center}
\includegraphics[width=\textwidth]{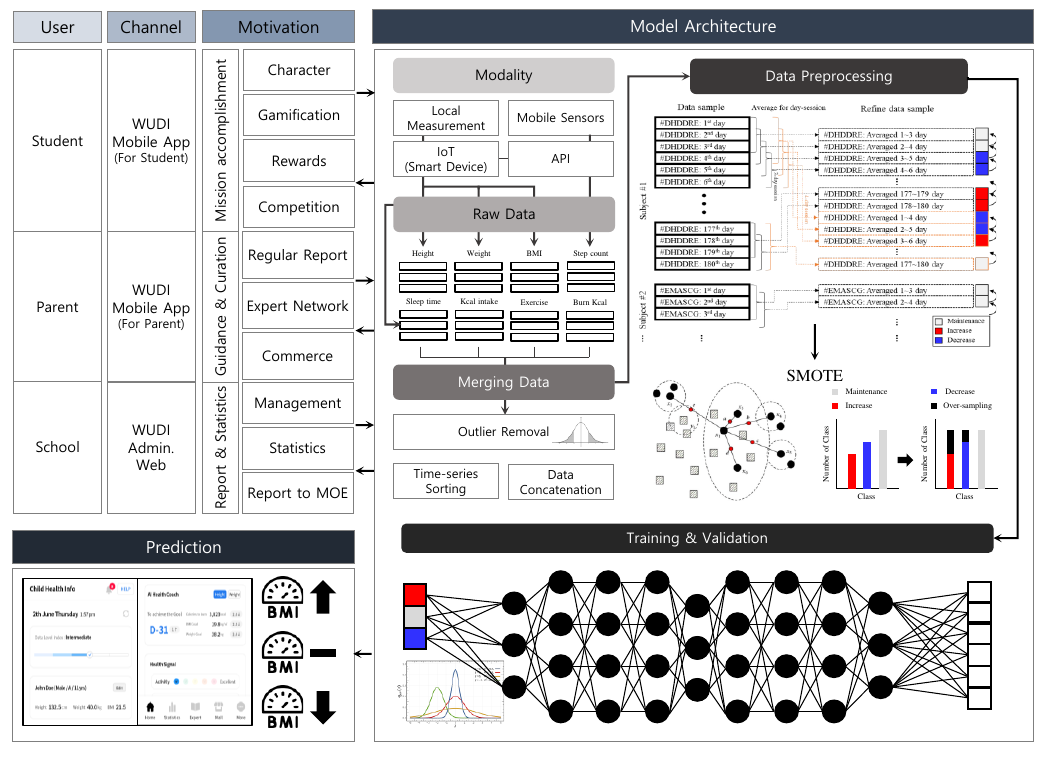}
\caption{Overall system framework for obesity prediction for adolescents. DeepHealthNet was proposed for training and evaluating the prediction. It comprises the outlier removal step, data preprocessing step for transforming the health data, application of the SMOTE step for augmenting the data, training, and validation step, and prediction step.}
\end{center}
\end{figure*}

WUDI! is a gamified mobile application that allows the student participants to choose a character as an avatar, customize the avatar with items purchasable with game coins, and complete missions and mini-games to gain points to level up the avatar and coins to purchase game items and lottery tickets for the monthly draw. Missions consisted of three parts: ``Play Well,'' ``Sleep Well,'' and ``Eat Well.'' Play Well tracked the activity of users, either by acquiring data from the smartwatch when they wore it, or obtaining the GPS data, moving distance, and moving minutes from the sensors on board the mobile phone when they did not wear the smartwatch. Sleep Well tracked the sleep of the users, either by acquiring data from the smartwatch when they wore it during sleep or obtaining the smartwatch usage information from the mobile phone with the log record module to know when the usage ceased at night and reactivated in the morning. Eat Well tracked the nutrition intake, asking the users to take a picture of food when they had a meal, and then analyzing the picture to understand the contents and caloric information of the meal via Vision AI analysis.

Each mission was assessed daily, and there were rewards on mission completion: points to level up and coins to purchase items. Mini-games were designed for user motivation, allowing them to play a limited number of rounds per day as a reward for daily mission completion in terms of nutrition (Eat Well) and activity (Play Well). Additionally, there were one-on-one competitions and a guild system, intended to motivate the application use by developing a sense of competition and bonding with each other. A contender can randomly choose another user to participate in the one-on-one competition to see who burned more calories within a given period. A guild could be formed with classmates, and there was a ranking board for the guild and individual users so that users could compete for individual ranks and guild ranks. In an approximately three-minute tutorial, participants completed the installation and registration of the two mobile applications, WUDI! and Samsung Health, and learned the proper usage of WUDI!.


The experimental period was eight weeks: four weeks during summer vacation and the remaining during the semester. The experiment was initially designed to compare the difference in overall measurements between two periods. After the initial body measurements, 321 participants started using WUDI! on their own mobile phones from the 4$^{th}$ week of July 2022. During the experiment, participants were free to use WUDI! voluntarily. However, there were two interventions from Injewelme Co., Ltd.. First, there were two monthly draws to award gifts based on the survey results of the participants. Second, events were randomly scheduled during the 3$^{rd}$ and 7$^{th}$ weeks to double the rewards awarded, such as points and coins, to sustain the high usage rate of WUDI! during the experiment period. Participants were free to undergo body measurements after the experiment. Accordingly, 187 participants completed both body measurements, but there were some participants who continued to use WUDI! without undergoing the second body measurement. The example data configuration from participants is represented in Table I.

\subsection{Proposed Deep Learning Framework}
Preprocessing the raw data is essential to ensure that the data are cleaned, formatted, and transformed into a suitable format that can be effectively used for ML models (Fig. 2). The characteristics used were height, weight, step count, burned calories, calorie intake, and total sleep time (Table I). In cases where height and weight were recorded multiple times in a day, the last day record was used. The data was then preprocessed into a time-series data format with multiple features, based on individuals and dates. However, this resulted in a significant number of missing values. All features were not recorded on the same date; thus, daily records were crucial to preserve the characteristics of time-series data. Therefore, instead of deleting rows with missing values, we chose to supplement missing values with the average of the nearest one or two values (Fig. 2).

To perform labeling, we conducted a preprocessing step that reduced n days' worth of data for an individual to a single row by using the average value of n days. This step involved two elements: how many days' worth of data are being reduced, and the length of the gap between the start dates of each bundle. For example, if we reduce 10 days of data by averaging every two days with two gaps, the data will be condensed into five data points. However, we only selected a gap length that allowed us to reduce all of the data up to the last day. For instance, if we reduce 10 days' worth of data into three bundles of three days with three gaps, the data on the 10$^{th}$ day cannot be condensed and hence are not included. The equation for calculating the bundle number is as follows. \textit{D} is the total number of record dates, \textit{t} is how many days' worth of data are being reduced, and \textit{m} is the length of the gap between the start days of each bundle. Only the integer result of the function is allowed.

\begin{align} 
Bundles=\frac{D-(n-1)}{m}
\end{align} 

The labels consisted of three categories: weight maintenance, weight gain, and weight loss. The first set of data was labeled as weight maintenance by default due to the absence of a comparison group. In most cases, there was no significant weight change between the previous and current groups when labeled using this method. Therefore, we used data augmentation using the synthetic minority oversampling technique (SMOTE) to solve the imbalance of the data. SMOTE is a popular data augmentation technique used to balance a class distribution by generating synthetic samples of the minority class \cite{chawla2002smote, elreedy2023theoretical}. It calculates the difference vector between a minority class sample and its nearest neighbor and generates new data points by scaling the difference vector by a random ratio. \textit{$x_{0}$} represents one of the candidates for integration as a minority class through SMOTE. \textit{$I_{B(x_{0},r)}$} represents the coverage of the minority class within a range with a radius of $r$, centered at \textit{$x_{0}$}. \textit{$pX(x)$} represents the original probability density of the minority class.
\begin{align} 
 I_{B(x_{0},r)}=\int_{B(x_{0},r)}pX(x)dx
\end{align} 
\textit{$z$}, the newly generated point, can be obtained by adding a uniform random variable \textit{$w$} multiplied by the vector difference between \textit{$x_k$} (a neighboring point) and \textit{$x_0$}:
\begin{align} 
\textrm{z}=(1-\textrm{w})x_{0}+\textrm{w}x_{k}
\end{align} 
The expression for the density function of point \textit{$z$} can be represented as follows. \textit{N} and \textit{K} represent the number of minority class samples and neighboring samples, respectively.
\begin{multline} 
p_{\textrm{Z}}(\textrm{z}) = \\
(N-K)\binom{N-1}{K} \int_{x}p_{X}(x)\int_{r=\left\|\textrm{z}-x \right\|}^{\infty}p_{X} \left ( x+\frac{(z-x)r}{\left\|\textrm{z}-x \right\|}\right ) \\
\times \left ( \frac{r^{d-2}}{\left\|\textrm{z}-x \right\|^{d-1}}\right ) B\left (1-I_{B(x,r)};N-K-1,K  \right )drdx
\end{multline}

The proposed architecture of the deep learning model consists of a neural network with multiple hidden layers to effectively capture complex patterns in the input data (Algorithm 1). The input layer, which is determined by the specific dimensions of the input data, receives the data for processing by the neural network. The first hidden layer comprises 128 densely connected nodes, where each node receives input from all nodes in the previous layer. These nodes apply their individual weights and biases to the received inputs, allowing them to learn and contribute to the representations of the network.

The second hidden layer consists of 256 densely connected nodes, mirroring the connectivity pattern of the previous layer. Each node in this layer receives inputs from all the nodes in the preceding layer and performs its own computations to extract higher-level features. This hierarchical structure enables the network to learn increasingly abstract representations as information flows through the layers.

\let\oldnl\nl 
\newcommand{\nonl}{\renewcommand{\nl}{\let\nl\oldnl}}
\SetKwInput{KwInput}{Input}              
\SetKwInput{KwOutput}{Output}
\SetKwInput{KwStep}{Step 1} \SetKwInput{KwStepp}{Step 2} \SetKwInput{KwSteppp}{Step 3} \SetKwInput{KwStepppp}{Step 4}

\begin{algorithm}[!t]
\linespread{1.0}
\footnotesize
\caption{Training procedure of DeepHealthNet}
\SetAlgoLined

\nonl $\bullet$ \KwIn{Training raw data} 
\nonl \ \ $X$ = \{${x}_{i}\}_{i=1}^{D}$, \{${x}_{i}\}\in\mathbb{R}^{P\times H}$: Training data for daily health, where $D$ is the total number of days, $P$ is the number of participants and $H$ is the number of health parameters (i.e., height, weight, BMI, step count, sleep time, kcal intake, exercise, burned kcal) \\
\nonl \ \ $\Omega$ = \{${O}_{i}\}_{i=1}^{D}$: Class labels, where ${O}_{i} \in \{Increase, Maintenance, Decrease\}$ and $D$ is the total number of days \\

\nonl $\bullet$ \KwOutput{Trained DeepHealthNet}
\nonl \hrulefill \\
\nonl $\bullet$ \KwStep{Preprocessing the input data}
\ \ Input $X_{bin}$: Merging data after the outlier removal\\
\ \ Transform n days' worth of data for an individual to a single row by using the average value of n days \\
\ \ Calculate the bundle features using Eq. (1) from the input data\\
\ \ Assign the class label according to the features of each bundle \\
\ \ Augment the data using SMOTE (Eq. (2)) \\
\ \ Output $X_{bin}$: Preprocessed data with class labels \\

\nonl $\bullet$ \KwStepp{Training the network}
\ \ Input $X_{bin}$: a set of preprocessed data\\
\ \ Input $\Omega$ = \{${O}_{tr}\}_{tr=1}^{D}$: multiclass labels, where ${O}_{tr} \in \{Increase, Maintenance, Decrease\}$, $D$ is the total number of days\\
\ \ The network parameters are initialized to random values for multiclass labels \\
\ \ Calculate feature maps extracted using Eq. (3)--Eq. (4)\\
\ \ Generate the loss value using Eq. (5)--Eq. (9) \\
\ \ Output $X_{N}$: Weights and loss values (multiclass) \\

\nonl $\bullet$ \KwSteppp{Fine-tune parameters}
\ \ Minimizing loss values by tuning the network parameters \\
\end{algorithm}


The third hidden layer consists of 128 densely connected nodes. Similarly to the previous layers, each node in this layer receives inputs from all nodes in the preceding layer. This layer further refines the learned features and contributes to the network's overall understanding of the input data. The output layer, which depends on the specific task at hand, generates the final outputs of the network. In this case, the number of nodes in the output layer corresponds to the three classes under consideration, namely ``Increase,'' ``Maintenance,'' and ``Decrease.'' Each node in the output layer represents the likelihood or probability of the input belonging to its corresponding class. For activation functions, the rectified linear unit (ReLU) function \cite{li2017convergence} is commonly used in the dense layers, including both the hidden and output layers. ReLU introduces non-linearity into the network, allowing it to learn complex relationships and adapt to various data patterns. However, depending on the requirements of the problem, other suitable activation functions can be utilized. The equation for the ReLU activation function is as follows:
\begin{align} 
& f(x) = \left \{ \begin{array}{rcl}
0 & \mbox{for} & x < 0\\ 
x & \mbox{for} & x \ge 0\end{array} \right. \\
& f(x) = max(0,x)
\end{align} 

The proposed model employs the cross-entropy loss function, which is widely used for multiclass classification problems \cite{zhang2018generalized}. Cross-entropy loss measures the dissimilarity between the predicted probabilities and the true class labels, guiding the network to minimize this discrepancy during training. By optimizing cross-entropy loss, the model aims to improve its performance for the prediction of obesity rates. The equation for the cross-entropy loss function is expressed as follows: y represents the true one-hot encoded label vector and $\hat{y}$ represents the predicted probability distribution between classes. \textit{p} is the predicted probability observation of the class.
\begin{align}
& CL(y, \hat{y}) = -(y\log(p) + (1 - y)\log(1 - p))\\
& CL(y, \hat{y}) = -\Sigma(y\log(\hat{y}))
\end{align}

In addition, the negative log-likelihood (NLL), minimum negative log-likelihood (MNL), and maximum likelihood estimation (MLE) are calculated simultaneously below. p(y) is a scalar rather than a vector. It is the value of the single dimension where the ground truth y lies. Thus, it is equivalent to cross-entropy. 
\begin{align}
& NLL(y) = -{\log(p(y))} \\
& MNL(y) = \min_{\theta} \sum_y {-\log(p(y;\theta))} \\
& MLE(y) = \max_{\theta} \prod_y p(y;\theta)
\end{align}


\subsection{Evaluation Methods \& Metrics}
\subsubsection{Comparison Models}
Several ML classification methods can be employed to predict obesity rates in adolescents using AI technology. The following are some commonly used classification methods for this task:

$\cdot$ Na{\"i}ve Bayes classifier (NB) \cite{reddy2022introduction}: The NB classifier is a probabilistic method that assumes that the features are independent of each other. It is effective for datasets with a large number of features and is computationally efficient.

$\cdot$ Regularized linear discriminant analysis (RLDA) \cite{zhao2014two}: RLDA is a statistical method used to find a linear combination of features that can effectively divide classes. It is effective for datasets with a small number of features and assumes that the data are normally distributed.

$\cdot$ Random forest (RF)\cite{breiman2001random}: This is an ensemble learning method that combines multiple decision trees to improve the accuracy of the classification. RF can handle a large number of features and is useful for feature selection. It handles imbalanced datasets effectively.

$\cdot$ Decision tree (DT) \cite{priyanka2020decision}: A DT is a tree-like structure that represents the decision-making process. It is a popular method for classification and can handle both numerical and categorical data. Decision trees are easy to interpret and visualize.

$\cdot$ Support vector machine (SVM) \cite{hearst1998support}: SVM is a supervised learning algorithm that uses a hyperplane to separate data points into different classes. It is effective for handling high-dimensional data and can handle nonlinearly separable data by using kernel functions \cite{noble2006support}.

$\cdot$ Long short-term memory (LSTM) \cite{hochreiter1997long}: LSTM is a type of recurrent neural network. It can handle sequential data, such as time-series data, and is effective for capturing long-term dependencies in the data.

\subsubsection{Performance Metrics}
Several evaluation metrics can be utilized to assess the performance of classification models. We used evaluation metrics, such as accuracy, precision, recall, and F1-score, to assess the performance of the model \cite{fawcett2006introduction, powers2020evaluation}. In addition, it is crucial to evaluate the models on different subsets of the data using 10-fold cross-validation to obtain reliable performance estimates. This approach helps obtain a more robust assessment of the performance and generalization ability of the model. In this study, the models were trained on 90\% of the data and tested on the remaining 10\% of the data.

\subsection{Statistical Analysis}
A statistical analysis was conducted to evaluate the performance of the ML models in predicting obesity rates in adolescents. To ensure the validity of the analysis, normality \cite{das2016brief} and homoskedasticity tests \cite{schultz1985levene} were performed on the data, considering the small sample size. The Shapiro-Wilk test was used to verify the normality of the data, and the results showed that the null hypothesis of normality was satisfied. Additionally, we confirmed homoskedasticity using Levene's test \cite{schultz1985levene} for each comparative group.

We conducted a paired $t$-test to compare the performances of the various ML models \cite{ross2017paired}. This statistical test allowed us to determine the statistical significance between the models and identify the most effective model for predicting obesity rates in adolescents. The results of the paired t-test were analyzed to provide insights into the performances of the ML models and to guide the selection of the best model for predicting obesity rates.

\begin{table*}[t!]
\centering
\small
\caption{Performance Evaluation of DeepHealthNet using Performance Metrics, such as Accuracy, F1-Score, Recall, and Precision}
\renewcommand{\arraystretch}{1.2}
\resizebox{\textwidth}{!}{%
\begin{tabular}{ccccccccccccccc}
\hline
 Models        & 1-fold               & 2-fold               & 3-fold               & 4-fold               & 5-fold               & 6-fold               & 7-fold               & 8-fold               & 9-fold               & 10-fold              & Accuracy        & F1-Score        & Recall          & Precision       \\ \hline
NB       & 0.3700          & 0.3704          & 0.3719          & 0.3724          & 0.3711          & 0.3721          & 0.3708          & 0.3706          & 0.3719          & 0.3727          & 0.3714          & 0.2536          & 0.5546          & 0.3714          \\
RLDA     & 0.4965          & 0.4969          & 0.4966          & 0.4957          & 0.4967          & 0.4978          & 0.4956          & 0.4967          & 0.4960          & 0.4953          & 0.4964          & 0.4939          & 0.4995          & 0.4964          \\
RF       & 0.5470          & 0.5473          & 0.5478          & 0.5474          & 0.5475          & 0.5479          & 0.5464          & 0.5478          & 0.5486          & 0.5492          & 0.5477          & 0.5365          & 0.5530          & 0.5477          \\
DT       & 0.6614          & 0.6618          & 0.6613          & 0.6614          & 0.6602          & 0.6605          & 0.6603          & 0.6596          & 0.6607          & 0.6613          & 0.6608          & 0.6616          & 0.6881          & 0.6608          \\
SVM      & 0.6851          & 0.6839          & 0.6841          & 0.6853          & 0.6850          & 0.6851          & 0.6847          & 0.6855          & 0.6834          & 0.6836          & 0.6846          & 0.6850          & 0.6891          & 0.6846          \\
LSTM      & 0.6948          & 0.6965          & 0.6984          & 0.7018          & 0.7015          & 0.7021          & 0.7021          & 0.7033          & 0.7024          & 0.7045          & 0.7008          & 0.7000          & 0.7037          & 0.7008          \\ 
DeepHealthNet & \textbf{0.8847} & \textbf{0.8854} & \textbf{0.8820} & \textbf{0.8836} & \textbf{0.8838} & \textbf{0.8824} & \textbf{0.8833} & \textbf{0.8825} & \textbf{0.8851} & \textbf{0.8842} & \textbf{0.8837} & \textbf{0.8797} & \textbf{0.8958} & \textbf{0.8837} \\ \hline
\end{tabular}}
\end{table*}

\begin{table}[t!]
\centering
\caption{Statistical Analysis of Differences between the Proposed and Compared Models in terms of Grand-Average Predicted Performances (Accuracy, F1-Score, Recall, Precision) }
\renewcommand{\arraystretch}{1.1}
\resizebox{\columnwidth}{!}{%
\begin{tabular}{rcccc}
\hline
\multicolumn{1}{l}{Comparison models} & Accuracy          & F1-Score          & Recall            & Precision         \\ \hline
NB vs. DeepHealthNet                      & $p$\textless{}0.001 & $p$\textless{}0.001 & $p$\textless{}0.001 & $p$\textless{}0.001 \\
RLDA vs. DeepHealthNet                       & $p$\textless{}0.001 & $p$\textless{}0.001 & $p$\textless{}0.001 & $p$\textless{}0.001 \\
RF vs. DeepHealthNet                     & $p$\textless{}0.001 & $p$\textless{}0.001 & $p$\textless{}0.001 & $p$\textless{}0.001 \\
DT vs. DeepHealthNet                       & $p$\textless{}0.001 & $p$\textless{}0.001 & $p$\textless{}0.001 & $p$\textless{}0.001 \\
SVM vs. DeepHealthNet                       & $p$\textless{}0.001 & $p$\textless{}0.001 & $p$\textless{}0.001 & $p$\textless{}0.001 \\
LSTM vs. DeepHealthNet                      & $p$\textless{}0.001 & $p$\textless{}0.05  & $p$\textless{}0.05  & $p$\textless{}0.05  \\ \hline
\end{tabular}}
\end{table}

\section{Experimental Results}
\subsection{Predicted Performances using Proposed and Comparison Models}
The experimental results presented in Table II demonstrate the superior performance of the proposed deep learning framework. It achieved an impressive average accuracy of 0.8837, outperforming all the compared models. The F1-score, which measures the balance between precision and recall, was also high at 0.8797. The recall value, which indicates the proportion of true positives identified, was 0.8958, whereas the precision value, representing the accuracy of positive predictions, matched the overall accuracy at 0.8837. Compared with the other models, the LSTM model exhibited a relatively high accuracy of 0.7008. Among the traditional ML classifiers, SVM demonstrated a commendable performance with an accuracy of 0.6846. In contrast, the NB model exhibited the lowest accuracy of 0.3727, which is comparable to random chance accuracy. Therefore, DeepHealthNet significantly outperformed all models compared in terms of accuracy, F1-score, recall, and precision. Across different folds, the standard deviation remained consistently below 0.02 for all models. Thus, there was no significant variance or instability during the training process. The performance metrics were consistently distributed, suggesting the robustness and reliability of the models throughout the training phase. Overall, the experimental results confirmed the effectiveness of DeepHealthNet, which consistently demonstrated superior accuracy and outperformed the models compared in terms of various performance measures.

Additionally, the results of the statistical analysis presented in Table III indicate the differences between the proposed deep learning framework and the comparison models in terms of various performance metrics. The obtained \textit{p}-values indicate the level of statistical significance and help assess whether the observed differences are coincidental or truly meaningful. As the \textit{p}-values were consistently less than 0.001 for accuracy, F1-score, recall, and precision, there is a highly significant statistical difference between the proposed deep learning framework and the comparison models in terms of these measures. However, when comparing the proposed deep learning framework with the LSTM model in terms of recall and precision, the obtained \textit{p}-value was less than 0.05, indicating a statistically significant difference. Although the proposed model generally outperformed the LSTM, this finding suggests that there are certain scenarios in which the LSTM might exhibit comparable performance in terms of recall and precision. 

\subsection{Performance Measurement using Confusion Matrix}
The study performed a detailed analysis of the confusion matrix for the proposed models to analyze their performance for each class using k-fold datasets, as illustrated in Fig. 3. The average accuracy of the proposed model was determined to be 0.8837, with the ``Decrease'' class exhibiting the highest true positive value at 0.9905. The ``Increase'' and ``Maintenance'' classes showed similar true positive rates (TPRs) of 0.9204 and 0.9720, respectively. Although there were some variations in performance between different classes, overall performance did not show significant differences. Thus, the proposed model was effective in predicting obesity rates in adolescents, demonstrating comparable performance across all the classes.

Among the models compared, the NB model achieved the highest TPR of 0.9171 for the ``Increase'' class. For the ``Maintenance'' and ``Decrease'' classes, the LSTM model exhibited the highest TPR values of 0.7765 and 0.9203, respectively. These findings suggest that the NB model correctly identified instances of increased obesity rates, whereas the LSTM model recognized instances of maintaining weight or decreasing obesity levels. The high accuracy observed in this experiment can be attributed to the availability of sufficient existing data to measure obesity in adolescents over a short period and providing valuable feedback, such as appropriate exercise recommendations, for managing obesity in this population. Consequently, the proposed model provided accurate predictions of obesity rates in adolescents, exhibiting high precision.

\subsection{Comparison of Predicted Performances between Boy and Girl Groups}
Table IV presents the results of predicting obesity by dividing the dataset into the boy and girl groups. The study aimed to explore whether there are differences in the manifestation of obesity between boys and girls and whether it would be more effective to analyze the prediction performance separately for each group. The proposed deep learning framework was utilized to assess the performance in predicting obesity in both groups. The results revealed that the obesity prediction performance achieved by the proposed model was 0.9320 for the boy group and 0.9163 for the girl group. Both groups exhibited similar performance, with no significant differences observed. The F1-scores for the boy and girl groups were 0.9318 and 0.9155, respectively, further indicating a comparable performance between the genders. Notably, the prediction performance in the boy group was approximately 2\% higher than that in the girl group. The proposed deep learning framework demonstrated the highest predictive performance for obesity, regardless of gender. Conversely, the compared models exhibited different levels of performance. Specifically, SVM showed the highest performance (0.7965) in the boy group, while LSTM (0.7675) outperformed the other models in the girl group. 

\begin{figure}[t!]
\begin{center}
\includegraphics[width=0.7\columnwidth]{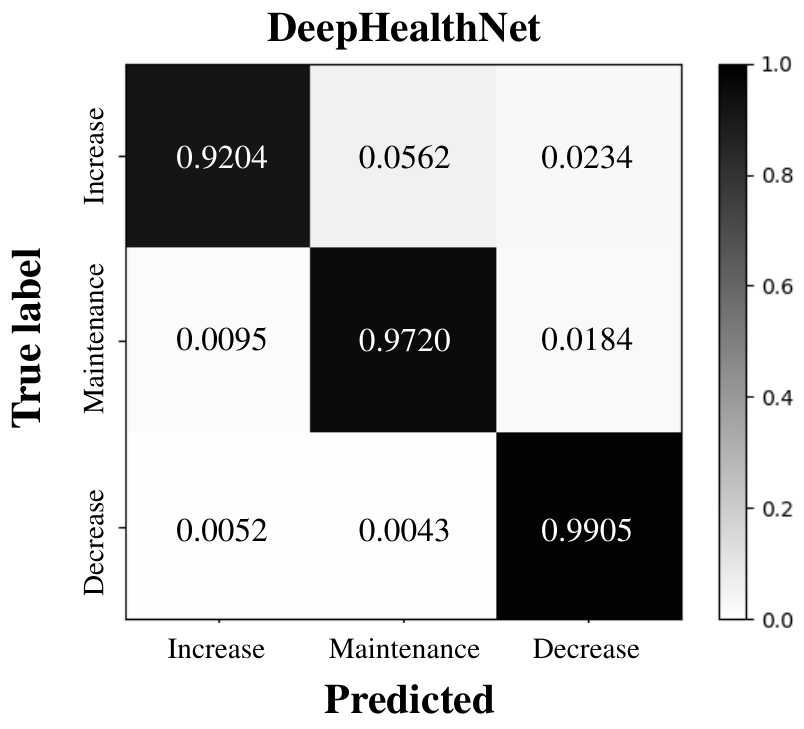}
\caption{Confusion matrices of each class (Increase, Maintenance, and Decrease) across all the participants using DeepHealthNet}
\end{center}
\end{figure}
\begin{table*}[t!]
\centering
\tiny
\caption{Performance Evaluation of Boy and Girl Groups using Performance Metrics, such as Accuracy, F1-Score, Recall, and Precision}
\renewcommand{\arraystretch}{1.0}
\resizebox{\textwidth}{!}{%
\begin{tabular}{cccccccccc}
\hline
\multicolumn{5}{c}{Boy   group}                                                  & \multicolumn{5}{c}{Girl   group}                                                 \\ \hline
Models   & Accuracy        & F1-Score        & Recall          & Precision       & Models   & Accuracy        & F1-Score        & Recall          & Precision       \\ \hline
NB       & 0.3779          & 0.2612          & 0.4822          & 0.3779          & NB       & 0.4076          & 0.3297          & 0.5216          & 0.4076          \\
RLDA     & 0.5089          & 0.5043          & 0.5156          & 0.5089          & RLDA     & 0.5087          & 0.5082          & 0.5099          & 0.5087          \\
RF       & 0.6459          & 0.6472          & 0.6556          & 0.6459          & RF       & 0.5840          & 0.5815          & 0.6005          & 0.5840          \\
DT       & 0.7787          & 0.7787          & 0.7991          & 0.7787          & DT       & 0.6914          & 0.6931          & 0.7302          & 0.6914          \\
SVM      & 0.7965          & 0.7964          & 0.7993          & 0.7965          & SVM      & 0.6909          & 0.6906          & 0.6987          & 0.6909          \\
LSTM     & 0.6493          & 0.6491          & 0.6500          & 0.6493          & LSTM     & 0.7675          & 0.7664          & 0.7678          & 0.7675          \\
DeepHealthNet & \textbf{0.9320} & \textbf{0.9318} & \textbf{0.9346} & \textbf{0.9320} & DeepHealthNet & \textbf{0.9163} & \textbf{0.9155} & \textbf{0.9206} & \textbf{0.9163} \\ \hline
\end{tabular}}
\end{table*}
\begin{table*}[t!]
\centering
\Huge
\caption{Statistical Analysis of Differences between Proposed and Compared Models in terms of Grand-average Predicted Performances (Accuracy, F1-Score, Recall, and Precision) }
\renewcommand{\arraystretch}{1.1}
\resizebox{\textwidth}{!}{%
\begin{tabular}{rccccccccccccc}
\hline
\multicolumn{1}{c}{}                   & \multicolumn{4}{c}{Within boy group}                                          & \multicolumn{4}{c}{Within girl group}                                         &                                            & \multicolumn{4}{c}{between boy and girl group}                                \\ \hline
\multicolumn{1}{c|}{Comparative models} & Accuracy          & F1-Score          & Recall            & Precision         & Accuracy          & F1-Score          & Recall            & Precision         & \multicolumn{1}{c|}{Comparative models}     & Accuracy          & F1-Score          & Recall            & Precision         \\ \hline
\multicolumn{1}{r|}{NB vs. DeepHealthNet}  & $p$\textless{}0.001 & $p$\textless{}0.001 & $p$\textless{}0.001 & $p$\textless{}0.001 & $p$\textless{}0.001 & $p$\textless{}0.001 & $p$\textless{}0.001 & $p$\textless{}0.001 & \multicolumn{1}{c|}{NB vs. NB}           & $p$\textgreater{}0.01 & $p$\textgreater{}0.01 & $p$\textgreater{}0.01 & $p$\textgreater{}0.01 \\
\multicolumn{1}{r|}{RLDA vs. DeepHealthNet}   & $p$\textless{}0.001 & $p$\textless{}0.001 & $p$\textless{}0.001 & $p$\textless{}0.001 & $p$\textless{}0.001 & $p$\textless{}0.001 & $p$\textless{}0.001 & $p$\textless{}0.001 & \multicolumn{1}{c|}{RLDA vs. RLDA}             & $p$\textgreater{}0.01 & $p$\textgreater{}0.01 & $p$\textgreater{}0.01 & $p$\textgreater{}0.01 \\
\multicolumn{1}{r|}{RF vs. DeepHealthNet} & $p$\textless{}0.001 & $p$\textless{}0.001 & $p$\textless{}0.001 & $p$\textless{}0.001 & $p$\textless{}0.001 & $p$\textless{}0.001 & $p$\textless{}0.001 & $p$\textless{}0.001 & \multicolumn{1}{c|}{RF vs. RF}         & $p$\textgreater{}0.01 & $p$\textgreater{}0.01 & $p$\textgreater{}0.01 & $p$\textgreater{}0.01 \\
\multicolumn{1}{r|}{DT vs. DeepHealthNet}   & $p$\textless{}0.001 & $p$\textless{}0.001 & $p$\textless{}0.001 & $p$\textless{}0.001 & $p$\textless{}0.001 & $p$\textless{}0.001 & $p$\textless{}0.001 & $p$\textless{}0.001 & \multicolumn{1}{c|}{DT vs. DT}             & $p$\textless{}0.005 & $p$\textless{}0.005 & $p$\textgreater{}0.01 & $p$\textless{}0.005 \\
\multicolumn{1}{r|}{SVM vs. DeepHealthNet}   & $p$\textless{}0.001 & $p$\textless{}0.001 & $p$\textless{}0.001 & $p$\textless{}0.001 & $p$\textless{}0.001 & $p$\textless{}0.001 & $p$\textless{}0.001 & $p$\textless{}0.001 & \multicolumn{1}{c|}{SVM vs. SVM}             & $p$\textless{}0.001 & $p$\textless{}0.001 & $p$\textless{}0.001 & $p$\textless{}0.001 \\
\multicolumn{1}{r|}{LSTM vs. DeepHealthNet} & $p$\textless{}0.001 & $p$\textless{}0.05  & $p$\textless{}0.05  & $p$\textless{}0.05  & $p$\textless{}0.001 & $p$\textless{}0.05  & $p$\textless{}0.05  & $p$\textless{}0.05  & \multicolumn{1}{c|}{LSTM vs. LSTM}         & $p$\textless{}0.001 & $p$\textless{}0.001 & $p$\textless{}0.001 & $p$\textless{}0.001 \\
\multicolumn{1}{c|}{}                  &                   &                   &                   &                   &                   &                   &                   &                   & \multicolumn{1}{c|}{DeepHealthNet vs. DeepHealthNet} & $p$\textgreater{}0.01 & $p$\textgreater{}0.01  & $p$\textgreater{}0.01  & $p$\textgreater{}0.01  \\ \hline
\end{tabular}}
\end{table*}
Table V presents an analysis of the statistically significant differences within each gender group for the compared models and the proposed deep learning framework. The aim of this analysis was to examine whether there were differences in the obesity prediction performance between the boy and girl groups and to determine if the proposed model outperformed the compared models within each group. Statistical analysis revealed that the obesity prediction performance of the proposed deep learning framework was more statistically significant (\textit{p}$<$0.001) compared to the other general models, even when considering the boy and girl groups separately. This highlights the superiority of the proposed model in accurately predicting obesity in both genders.

Furthermore, the results presented in Table V can be used to examine whether there were significant differences in the obesity prediction performance between the boys and girls groups for each of the models compared. The results showed that for models such as DT, SVM, and LSTM, there was a statistically significant difference in performance between the boy and girl groups (\textit{p}$<$0.005 and \textit{p}$<$0.001). On the other hand, while there were no statistically significant differences for the other compared models (i.e., NB, RLDA, RF, and the proposed model), their obesity prediction performance was significantly lower. Overall, the proposed deep learning framework demonstrated a higher obesity prediction performance than the other models, without significant differences between the gender groups. This indicates that the proposed model can effectively address the imbalanced data issue according to gender. The preprocessing module employed in the framework helps mitigate this issue, ensuring reliable and accurate predictions regardless of the gender distribution in the dataset.


\subsection{Performance Comparison using Each Day Session}
Fig. 4(a) illustrates the obesity prediction performance for each day session. The graph illustrates the relationship between the day session and the corresponding predictive performance of the proposed model. As the day sessions progressed, the predictive performance of obesity gradually improved. According to the proposed model, around the 60th-day session, the performance reached saturation, indicating that the model had learned and captured the underlying patterns in the data. In particular, the highest prediction performance of obesity was achieved in the 162nd-day session, recorded as 0.8848. However, the performance gradually decreased after the 170th-day session.

In contrast, the prediction performance of the other compared model groups showed different patterns. For instance, in the case of LSTM, the highest performance was observed in the 106th-day session, but subsequently, the performance exhibited fluctuations with alternating increases and decreases. These findings suggest that the proposed model can consistently predict the obesity levels of the target population for future time intervals, typically spanning approximately 5 to 6 months. However, the performances of the other compared models showed uncertainty in determining the appropriate time duration for collecting new datasets to achieve accurate predictions. The performances of the compared models varied, and the required period of data collection to achieve reliable predictions is unclear. These results emphasized the effectiveness and long-term predictive capabilities of the proposed model in predicting obesity levels over an extended period.
\begin{figure*}[t!]
\begin{center}
\includegraphics[width=0.85\textwidth]{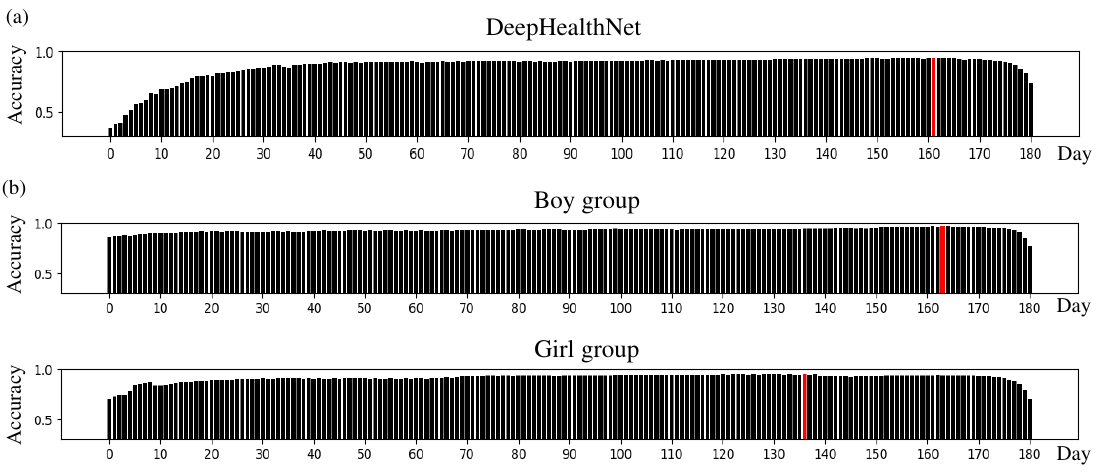}
\caption{Comparison of classification accuracy per nth-day sessions. (a) Classification accuracy per day session using DeepHealthNet and the compared models. The red bar indicates the highest accuracy across all the day sessions. (b) Comparison of classification accuracy using DeepHealthNet by gender. In the 164th-day session, the boy group achieved the highest accuracy, and the girl group exhibited the highest accuracy in the 137th-day session.}
\end{center}
\end{figure*}
Fig. 4 (b) shows the temporal evolution of obesity prediction performance using the proposed model, specifically for each gender group. The patterns depicted in Fig. 4(b) show the trajectory of obesity prediction performance for both genders. The graph displays the variations in performance over several days for boys and girls. In the 164th-day session, the boy group achieved the highest result (0.9320), similar to the 162nd-day session, which encompassed the entire data sample. In contrast, the group of girls exhibited the highest performance (0.9163) in the 137th-day session. The day session representing the peak performance between the two gender groups had a one-month difference. This suggests that significant physical changes, including BMI, occurred over approximately four months for the girl group. In contrast, six months of daily health data was required to accurately predict BMI for the boy group, indicating a slower pace of physical change compared to the girl group. These contrasting trends in physical changes between the male and female groups were reflected in the deep learning model. The findings confirmed that the data obtained were also considered, given the faster rate of physical change in the girl group compared to the boy group.



\section{Discussions}
The results of the study demonstrated the excellent performance of the proposed deep learning framework for predicting obesity levels in adolescents. The experimental findings indicated that the proposed model outperformed all the compared models in terms of accuracy, F1-score, recall, and precision. The average accuracy achieved by the proposed model was 0.8837, surpassing the performance of all the other models. The F1-score, which measures the balance between precision and recall, was also high at 0.8797. The recall value, which represents the proportion of true positives identified, was 0.8958, and the precision value, indicating the accuracy of positive predictions, matched the overall accuracy at 0.8837. Furthermore, statistical analysis indicated the significance of the differences between the proposed deep learning framework and the models compared in terms of various performance metrics. The obtained $p$-values were consistently below 0.001, indicating a highly significant statistical difference in favor of the proposed model. 

The study also investigated the model's prediction performance by dividing the dataset into boy and girl groups. The results indicated that the model achieved similar obesity prediction performances in both groups, with no significant difference observed. This finding suggested that the proposed deep learning framework was effective in predicting obesity rates regardless of gender, demonstrating its robustness and generalizability. The ability of the model to perform well in both boy and girl groups is significant because it demonstrates that the model is not biased toward a particular gender. This suggests that the features and patterns captured by the model are relevant and applicable to both genders for predicting obesity levels. This finding has practical implications as it indicates that the model can be utilized in diverse populations without the need for gender-specific modifications or adjustments.

Furthermore, the statistical analysis conducted within each gender group reinforced the superiority of the proposed model. Even when considering the boy and girl groups separately, the obesity prediction performance of the model was more statistically significant compared to that of the other models. This highlights its effectiveness in accurately predicting obesity levels in both boys and girls and provides additional evidence for its robustness. Overall, the prediction performance in separate boy and girl groups emphasized the effectiveness of the proposed deep learning framework in accurately predicting obesity levels regardless of gender. This strengthens the applicability and reliability of the model in diverse populations and contributes to its generalizability.

The visualization of the obesity prediction performance over time provided valuable insights into the long-term predictive capabilities of the proposed model. The graph demonstrated a gradual improvement in the model's performance as the duration increased, with the highest prediction performance achieved on the 162$^{nd}$ day session. This indicates that the proposed model could consistently forecast obesity levels over an extended period. The improvement in performance over time suggests that the model benefits from a longer duration of data collection and learning. As more data become available, the model can capture more patterns and relationships, leading to enhanced predictive accuracy. The saturation point was reached after approximately 60 days, which suggests that the model's performance plateaus and that collecting data beyond this point may not significantly improve its predictive capabilities. The compared models exhibited varying levels of performance highlighting the uncertainty associated with determining the optimal duration for collecting new datasets. In contrast, the proposed model demonstrated consistent and reliable long-term predictions, outperforming the compared models. The performance visualization of the model over time indicates that the proposed model can be utilized for long-term obesity prediction, offering reliable forecasts beyond shorter timeframes. This knowledge is valuable in decision-making processes related to interventions, public health policies, and resource allocation, as it allows for more accurate and informed predictions of obesity levels over extended periods.

Nevertheless, this study has some limitations. First, the study focused on a specific age group, and the generalizability of the proposed model to other age groups or populations needs further investigation. Second, the study utilized a specific dataset for training and evaluation, and the generalizability of the proposed model to other datasets should be explored. Additionally, the study did not consider certain factors that could influence obesity, such as socioeconomic status, dietary habits, or genetic factors. Incorporating these factors into the model should enhance its predictive capabilities. Furthermore, the study employed a deep learning framework, which may require substantial computational resources and expertise for implementation. The feasibility and practicality of the proposed model in real-world settings need further consideration. Finally, the study did not conduct a longitudinal analysis to assess the performance of the model over an extended period beyond the available data. In future studies, we look forward to investigating the stability and accuracy of the model over a more extended timeframe.

\section{Conclusion and Future Works}
This study proposed a deep learning framework for predicting obesity levels in adolescents. The proposed model demonstrated superior performance compared with other models, achieving high accuracy, F1-score, recall, and precision values. Statistical analysis confirmed the significant differences in favor of the proposed model. The effectiveness of the model was consistent across gender groups, highlighting its robustness. Visualizations proved the model's ability to provide reliable long-term predictions, outperforming the compared models. Hence, considering the limitations of this study and future research directions, we plan to enhance the generalizability of the model further, incorporate additional factors, and assess its performance over extended periods.


 \bibliographystyle{IEEEtran}
\bibliography{Reference}

\begin{thebibliography}{10}
\providecommand{\url}[1]{#1}
\csname url@samestyle\endcsname
\providecommand{\newblock}{\relax}
\providecommand{\bibinfo}[2]{#2}
\providecommand{\BIBentrySTDinterwordspacing}{\spaceskip=0pt\relax}
\providecommand{\BIBentryALTinterwordstretchfactor}{4}
\providecommand{\BIBentryALTinterwordspacing}{\spaceskip=\fontdimen2\font plus
\BIBentryALTinterwordstretchfactor\fontdimen3\font minus
  \fontdimen4\font\relax}
\providecommand{\BIBforeignlanguage}[2]{{%
\expandafter\ifx\csname l@#1\endcsname\relax
\typeout{** WARNING: IEEEtran.bst: No hyphenation pattern has been}%
\typeout{** loaded for the language `#1'. Using the pattern for}%
\typeout{** the default language instead.}%
\else
\language=\csname l@#1\endcsname
\fi
#2}}
\providecommand{\BIBdecl}{\relax}
\BIBdecl

\bibitem{abarca2017worldwide}
L.~Abarca-G{\'o}mez \emph{et~al.}, ``Worldwide trends in body-mass index,
  underweight, overweight, and obesity from 1975 to 2016: a pooled analysis of
  2416 population-based measurement studies in 128{\textperiodcentered} 9
  million children, adolescents, and adults,'' \emph{The lancet}, vol. 390, no.
  10113, pp. 2627--2642, 2017.

\bibitem{ferreras2023systematic}
A.~Ferreras, S.~Sumalla-Cano, R.~Mart{\'\i}nez-Licort, I.~El{\'\i}o,
  K.~Tutusaus, T.~Prola, J.~L. Vidal-Maz{\'o}n, B.~Sahelices, and I.~de~la
  Torre~D{\'\i}ez, ``Systematic review of machine learning applied to the
  prediction of obesity and overweight,'' \emph{Journal of Medical Systems},
  vol.~47, no.~1, pp. 1--11, 2023.

\bibitem{degregory2018review}
K.~DeGregory \emph{et~al.}, ``A review of machine learning in obesity,''
  \emph{Obesity Reviews}, vol.~19, no.~5, pp. 668--685, 2018.

\bibitem{caprio2020childhood}
S.~Caprio, N.~Santoro, and R.~Weiss, ``Childhood obesity and the associated
  rise in cardiometabolic complications,'' \emph{Nature metabolism}, vol.~2,
  no.~3, pp. 223--232, 2020.

\bibitem{ahmed2009childhood}
M.~L. Ahmed, K.~K. Ong, and D.~B. Dunger, ``Childhood obesity and the timing of
  puberty,'' \emph{Trends in Endocrinology \& Metabolism}, vol.~20, no.~5, pp.
  237--242, 2009.

\bibitem{hammond2019predicting}
R.~Hammond \emph{et~al.}, ``Predicting childhood obesity using electronic
  health records and publicly available data,'' \emph{PloS One}, vol.~14,
  no.~4, p. e0215571, 2019.

\bibitem{pang2019understanding}
X.~Pang, C.~B. Forrest, F.~L{\^e}-Scherban, and A.~J. Masino, ``Understanding
  early childhood obesity via interpretation of machine learning model
  predictions,'' in \emph{2019 18th IEEE International Conference On Machine
  Learning And Applications (ICMLA)}.\hskip 1em plus 0.5em minus 0.4em\relax
  IEEE, 2019, pp. 1438--1443.

\bibitem{torner2019multipurpose}
J.~Torner, S.~Skouras, J.~L. Molinuevo, J.~D. Gispert, and F.~Alpiste,
  ``Multipurpose virtual reality environment for biomedical and health
  applications,'' \emph{IEEE Transactions on Neural Systems and Rehabilitation
  Engineering}, vol.~27, no.~8, pp. 1511--1520, 2019.

\bibitem{saadeh2019patient}
W.~Saadeh, S.~A. Butt, and M.~A.~B. Altaf, ``A patient-specific single sensor
  {I}o{T}-based wearable fall prediction and detection system,'' \emph{IEEE
  Transactions on Neural Systems and Rehabilitation Engineering}, vol.~27,
  no.~5, pp. 995--1003, 2019.

\bibitem{ang2016eeg}
K.~K. Ang and C.~Guan, ``{EEG}-based strategies to detect motor imagery for
  control and rehabilitation,'' \emph{IEEE Transactions on Neural Systems and
  Rehabilitation Engineering}, vol.~25, no.~4, pp. 392--401, 2016.

\bibitem{jeong2020brain}
J.-H. Jeong, K.-H. Shim, D.-J. Kim, and S.-W. Lee, ``Brain-controlled robotic
  arm system based on multi-directional {CNN-BiLSTM} network using {EEG}
  signals,'' \emph{IEEE Transactions on Neural Systems and Rehabilitation
  Engineering}, vol.~28, no.~5, pp. 1226--1238, 2020.

\bibitem{zhao2022uda}
Z.~Zhao, F.~Zhou, K.~Xu, Z.~Zeng, C.~Guan, and S.~K. Zhou, ``{LE-UDA}:
  {L}abel-efficient unsupervised domain adaptation for medical image
  segmentation,'' \emph{IEEE Transactions on Medical Imaging}, 2022.

\bibitem{jeong2022real}
J.-H. Jeong, J.-H. Cho, B.-H. Lee, and S.-W. Lee, ``Real-time deep
  neurolinguistic learning enhances noninvasive neural language decoding for
  brain--machine interaction,'' \emph{IEEE Transactions on Cybernetics}, 2022.

\bibitem{yu2018artificial}
K.-H. Yu, A.~L. Beam, and I.~S. Kohane, ``Artificial intelligence in
  healthcare,'' \emph{Nature biomedical engineering}, vol.~2, no.~10, pp.
  719--731, 2018.

\bibitem{lee2021decoding}
M.~Lee, J.-H. Jeong, Y.-H. Kim, and S.-W. Lee, ``Decoding finger tapping with
  the affected hand in chronic stroke patients during motor imagery and
  execution,'' \emph{IEEE Transactions on Neural Systems and Rehabilitation
  Engineering}, vol.~29, pp. 1099--1109, 2021.

\bibitem{bastida2023promoting}
L.~Bastida \emph{et~al.}, ``Promoting obesity prevention and healthy habits in
  childhood: The ocariot experience,'' \emph{IEEE Journal of Translational
  Engineering in Health and Medicine}, vol.~11, pp. 261--270, 2023.

\bibitem{yang2020homecare}
G.~Yang, Z.~Pang, M.~J. Deen, M.~Dong, Y.-T. Zhang, N.~Lovell, and A.~M.
  Rahmani, ``Homecare robotic systems for healthcare 4.0: visions and enabling
  technologies,'' \emph{IEEE Journal of Biomedical and Health Informatics},
  vol.~24, no.~9, pp. 2535--2549, 2020.

\bibitem{huffman2010parenthood}
F.~G. Huffman, S.~Kanikireddy, and M.~Patel, ``Parenthood—a contributing
  factor to childhood obesity,'' \emph{International journal of environmental
  research and public health}, vol.~7, pp. 2800--2810, 2010.

\bibitem{mondal2023predicting}
P.~K. Mondal, K.~H. Foysal, B.~A. Norman, and L.~S. Gittner, ``Predicting
  childhood obesity based on single and multiple well-child visit data using
  machine learning classifiers,'' \emph{Sensors}, vol.~23, no.~2, p. 759, 2023.

\bibitem{gupta2022obesity}
M.~Gupta, T.-L.~T. Phan, H.~T. Bunnell, and R.~Beheshti, ``Obesity prediction
  with ehr data: A deep learning approach with interpretable elements,''
  \emph{ACM Transactions on Computing for Healthcare (HEALTH)}, vol.~3, no.~3,
  pp. 1--19, 2022.

\bibitem{cheng2022predicting}
E.~R. Cheng, R.~Steinhardt, and Z.~Ben~Miled, ``Predicting childhood obesity
  using machine learning: Practical considerations,'' \emph{BioMedInformatics},
  vol.~2, no.~1, pp. 184--203, 2022.

\bibitem{chawla2002smote}
N.~V. Chawla, K.~W. Bowyer, L.~O. Hall, and W.~P. Kegelmeyer, ``Smote:
  synthetic minority over-sampling technique,'' \emph{Journal of artificial
  intelligence research}, vol.~16, pp. 321--357, 2002.

\bibitem{elreedy2023theoretical}
D.~Elreedy, A.~F. Atiya, and F.~Kamalov, ``A theoretical distribution analysis
  of synthetic minority oversampling technique ({SMOTE}) for imbalanced
  learning,'' \emph{Machine Learning}, pp. 1--21, 2023.

\bibitem{li2017convergence}
Y.~Li and Y.~Yuan, ``Convergence analysis of two-layer neural networks with
  relu activation,'' \emph{Advances in neural information processing systems
  (NIPS)}, vol.~30, 2017.

\bibitem{zhang2018generalized}
Z.~Zhang and M.~Sabuncu, ``Generalized cross entropy loss for training deep
  neural networks with noisy labels,'' \emph{Advances in neural information
  processing systems}, vol.~31, 2018.

\bibitem{reddy2022introduction}
E.~M.~K. Reddy, A.~Gurrala, V.~B. Hasitha, and K.~V.~R. Kumar, ``Introduction
  to naive bayes and a review on its subtypes with applications,''
  \emph{Bayesian Reasoning and Gaussian Processes for Machine Learning
  Applications}, pp. 1--14, 2022.

\bibitem{zhao2014two}
J.~Zhao, L.~Shi, and J.~Zhu, ``Two-stage regularized linear discriminant
  analysis for 2-d data,'' \emph{IEEE Transactions on Neural Networks and
  Learning Systems}, vol.~26, no.~8, pp. 1669--1681, 2014.

\bibitem{breiman2001random}
L.~Breiman, ``Random forests,'' \emph{Machine learning}, vol.~45, pp. 5--32,
  2001.

\bibitem{priyanka2020decision}
Priyanka and D.~Kumar, ``Decision tree classifier: a detailed survey,''
  \emph{International Journal of Information and Decision Sciences}, vol.~12,
  no.~3, pp. 246--269, 2020.

\bibitem{hearst1998support}
M.~A. Hearst, S.~T. Dumais, E.~Osuna, J.~Platt, and B.~Scholkopf, ``Support
  vector machines,'' \emph{IEEE Intelligent Systems \& Their Applications},
  vol.~13, no.~4, pp. 18--28, 1998.

\bibitem{noble2006support}
W.~S. Noble, ``What is a support vector machine?'' \emph{Nature biotechnology},
  vol.~24, no.~12, pp. 1565--1567, 2006.

\bibitem{hochreiter1997long}
S.~Hochreiter and J.~Schmidhuber, ``Long short-term memory,'' \emph{Neural
  computation}, vol.~9, no.~8, pp. 1735--1780, 1997.

\bibitem{fawcett2006introduction}
T.~Fawcett, ``An introduction to {ROC} analysis,'' \emph{Pattern Recognit.
  Lett.}, vol.~27, no.~8, pp. 861--874, 2006.

\bibitem{powers2020evaluation}
D.~M. Powers, ``Evaluation: from precision, recall and f-measure to roc,
  informedness, markedness and correlation,'' \emph{arXiv preprint
  arXiv:2010.16061}, 2020.

\bibitem{das2016brief}
K.~R. Das and A.~Imon, ``A brief review of tests for normality,''
  \emph{American Journal of Theoretical and Applied Statistics}, vol.~5, no.~1,
  pp. 5--12, 2016.

\bibitem{schultz1985levene}
B.~B. Schultz, ``Levene's test for relative variation,'' \emph{Systematic
  Zoology}, vol.~34, no.~4, pp. 449--456, 1985.

\bibitem{ross2017paired}
A.~Ross, V.~L. Willson, A.~Ross, and V.~L. Willson, ``Paired samples t-test,''
  \emph{Basic and Advanced Statistical Tests: Writing Results Sections and
  Creating Tables and Figures}, pp. 17--19, 2017.

\end{thebibliography}

\end{document}

\begin{tabular}{ccccc}
\hline
\multicolumn{5}{c}{Boy   Group}                                                  \\ \hline
Models   & Accuracy        & F1-Score        & Recall          & Precision       \\ \hline
SVM      & 0.5645          & 0.5666          & 0.5774          & 0.5645          \\
RF       & 0.4918          & 0.4768          & 0.4994          & 0.4918          \\
RLDA     & 0.4560          & 0.4531          & 0.4554          & 0.4560          \\
DT       & 0.5372          & 0.5273          & 0.5781          & 0.5372          \\
NB       & 0.3944          & 0.3018          & 0.3332          & 0.3944          \\
LSTM     & 0.7794          & 0.7739          & 0.7851          & 0.7794          \\
Proposed & \textbf{0.8362} & \textbf{0.7913} & \textbf{0.7998} & \textbf{0.7959} \\ \hline
\multicolumn{5}{c}{}                                                             \\ \hline
\multicolumn{5}{c}{Girl Group}                                                   \\ \hline
Models   & Accuracy        & F1-Score        & Recall          & Precision       \\ \hline
SVM      & 0.5645          & 0.5666          & 0.5774          & 0.5645          \\
RF       & 0.4918          & 0.4768          & 0.4994          & 0.4918          \\
RLDA     & 0.4560          & 0.4531          & 0.4554          & 0.4560          \\
DT       & 0.5372          & 0.5273          & 0.5781          & 0.5372          \\
NB       & 0.3944          & 0.3018          & 0.3332          & 0.3944          \\
LSTM     & 0.7794          & 0.7739          & 0.7851          & 0.7794          \\
Proposed & \textbf{0.8362} & \textbf{0.7913} & \textbf{0.7998} & \textbf{0.7959} \\ \hline
\end{tabular}